\documentclass[lettersize,journal]{IEEEtran}
\usepackage{amsmath,amsfonts}
\usepackage{algorithmic}
\usepackage{amssymb}
\usepackage{algorithm}
\usepackage{array}
\usepackage[caption=false,font=normalsize,labelfont=sf,textfont=sf]{subfig}
\usepackage{textcomp}
\usepackage{stfloats}
\usepackage{url}
\usepackage{xcolor}
\usepackage {multirow}
\usepackage{textcomp}
\usepackage{booktabs}
\usepackage{verbatim}
\usepackage{graphicx}
\usepackage[utf8]{inputenc}
\usepackage{cite}
\usepackage[a4paper, left=1.8cm, right=1.8cm, top=1.9cm, bottom=1.7cm]{geometry}
\begin{document}

\title{Hierarchical Graph Feature Enhancement with Adaptive Frequency Modulation for Visual Recognition}

\author{Feiyue~Zhao and Zhichao~Zhang,~\IEEEmembership{Member,~IEEE}
\thanks{This work was supported in part by the Open Foundation of Hubei Key Laboratory of Applied Mathematics (Hubei University) under Grant HBAM202404; in part by the Foundation of Key Laboratory of System Control and Information Processing, Ministry of Education under Grant Scip20240121; in part by the Foundation of Key Laboratory of Computational Science and Application of Hainan Province under Grant JSKX202401; and in part by the Foundation of Key Laboratory of Numerical Simulation of Sichuan Provincial Universities under Grant KLNS--2024SZFZ005. \emph{(Corresponding author: Zhichao~Zhang.)}}
\thanks{Feiyue~Zhao is with the School of Mathematics and Statistics, the Center for Applied Mathematics of Jiangsu Province, and the Jiangsu International Joint Laboratory on System Modeling and Data Analysis, Nanjing University of Information Science and Technology, Nanjing 210044, China (e-mail: 202511150010@nuist.edu.cn).}
\thanks{Zhichao~Zhang is with the School of Mathematics and Statistics, Nanjing University of Information Science and Technology, Nanjing 210044, China, with the Hubei Key Laboratory of Applied Mathematics, Hubei University, Wuhan 430062, China, with the Key Laboratory of System Control and Information Processing, Ministry of Education, Shanghai Jiao Tong University, Shanghai 200240, China, with the Key Laboratory of Computational Science and Application of Hainan Province, Hainan Normal University, Haikou 571158, China, and also with the Key Laboratory of Numerical Simulation of Sichuan Provincial Universities, School of Mathematics and Information Sciences, Neijiang Normal University, Neijiang 641000, China (e-mail: zzc910731@163.com).}}

\markboth{Journal of \LaTeX\ Class Files,~Vol.~14, No.~8, August~2021}%
{Shell \MakeLowercase{\textit{et al.}}: A Sample Article Using IEEEtran.cls for IEEE Journals}


\maketitle

\begin{abstract}
Convolutional neural networks (CNNs) have demonstrated strong performance in visual recognition tasks, but their inherent reliance on regular grid structures limits their capacity to model complex topological relationships and non-local semantics within images. To address this limitation, we propose the hierarchical graph feature enhancement (HGFE), a novel framework that integrates graph-based reasoning into CNNs to enhance both structural awareness and feature representation. HGFE builds two complementary levels of graph structures: intra-window graph convolution to capture local spatial dependencies and inter-window supernode interactions to model global semantic relationships. Moreover, we introduce an adaptive frequency modulation module that dynamically balances low-frequency and high-frequency signal propagation, preserving critical edge and texture information while mitigating over-smoothing. The proposed HGFE module is lightweight, end-to-end trainable, and can be seamlessly integrated into standard CNN backbone networks. Extensive experiments on CIFAR-100 (classification), PASCAL VOC, and VisDrone (detection), as well as CrackSeg and CarParts (segmentation), validated the effectiveness of the HGFE in improving structural representation and enhancing overall recognition performance.
\end{abstract}


\section{Introduction}
\IEEEPARstart{R}{ecently}, convolutional neural networks (CNNs) have established themselves as the fundamental backbone in the field of computer vision \cite{lenet, alexnet, vgg, googlenet, resnet, densenet, efficientnet}, achieving remarkable success across a wide spectrum of image processing tasks, including image classification \cite{cls0, cls1, cls2, cls3}, object detection \cite{det0yoloword, det1, det2yolov10, det3yolov12}, and instance segmentation \cite{seg0, seg1, seg2, seg3}. By leveraging hierarchical feature extraction and localized convolutional kernels, CNNs are particularly effective in capturing spatially local patterns and texture information. This inductive bias enables CNNs to learn robust representations that generalize well in various visual recognition scenarios.

However, the fundamental design of CNNs imposes a key limitation: reliance on fixed-size local receptive fields and regular grid structures. Although deeper architectures or dilated convolutions can expand the effective receptive field, the intrinsic locality of convolutional operations restricts the ability of CNNs to model long-range dependencies and capture irregular or complex topological relationships present in visual data. This limitation is particularly critical for tasks requiring a holistic understanding of scene structure or contextual reasoning, such as dense prediction and object-level understanding.

To overcome this bottleneck, several architectural innovations have been proposed \cite{convtrans0, convtrans1, convtrans2, convtrans3}. Self-attention mechanisms \cite{att}, notably introduced by the vision transformer (ViT) \cite{vit}, offer a method for capturing global dependencies by modeling pairwise interactions across all spatial positions. Similarly, non-local neural networks provide global context modeling through affinity-based aggregation \cite{nonnet0, nonnet1, nonnet2, nonnet3}. Despite their success, these approaches often incur high computational and memory overheads, particularly for high-resolution inputs. Furthermore, transformer-based models typically require large-scale datasets and sophisticated training strategies to converge effectively.

In parallel, graph convolutional networks (GCNs) have emerged as a promising alternative for non-Euclidean representation learning \cite{gnn0, gnn1, gnn2, gat3, gnn4}, and are particularly well-suited for modeling relational structures and irregular spatial dependencies. By representing features as nodes and constructing edges based on semantic or spatial proximity, GCNs enable explicit message passing between nonlocal regions, thus providing a flexible and powerful tool for structural modeling. Recent efforts have incorporated GCNs into visual pipelines, achieving progress in tasks such as skeleton-based action recognition \cite{gcnvis0}, \cite{gcnvis1}, point-cloud segmentation \cite{gcnvis2}, \cite{gcnvis3}, and scene graph generation \cite{gcnvis4}. However, designing effective graph structures that simultaneously capture fine-grained local details and broader semantic contexts while remaining computationally efficient remains an open challenge.

Moreover, the spectral characteristics of graph-structured features are often underutilized in conventional models. Without careful control, GCNs tend to suffer from over-smoothing, where repeated propagation leads to feature homogenization and loss of discriminative detail. A more principled handling of spectral components, particularly adaptive control over low-frequency and high-frequency propagation, is crucial for maintaining both structural integrity and task relevance.
\begin{figure*}[!t]
\centering
\includegraphics[width=7.0in]{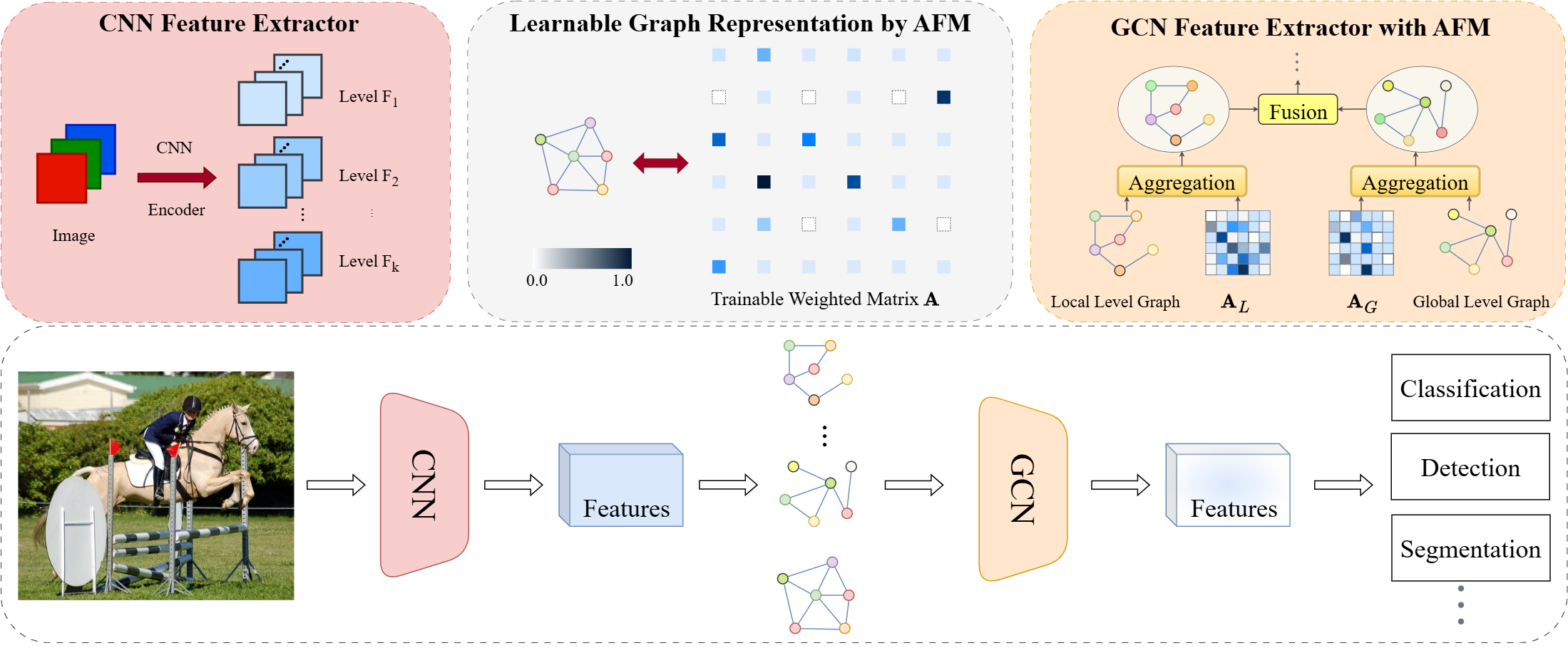}
\caption{Overall architecture of the proposed HGFE-enhanced vision backbone. The model integrates HGFE modules into a standard vision backbone, enabling adaptive frequency-aware feature propagation to improve spatial detail preservation and semantic representation across multiple stages.}
\label{fig:framework}
\end{figure*}

To address these limitations, we propose a novel hierarchical graph feature enhancement (HGFE) framework that integrates graph-based reasoning into standard CNN architectures to enhance their structural modeling capability and contextual representation. Specifically, as shown in Fig.~\ref{fig:framework}, our approach introduces a dual-level graph convolution strategy. At the local level, we constructed intra-window graphs that captured fine-grained spatial dependencies within localized regions. At the global level, we introduced inter-window supernode graphs that enabled efficient context aggregation across distant regions via compact supernode interactions.

To further improve the representational capacity of graph convolution, we introduce an adaptive frequency modulation (AFM) mechanism based on graph signal processing (GSP) theory. This module dynamically modulates the spectral response of the graph filters by selectively balancing the low-frequency (smooth) and high-frequency (detail-rich) components according to the input feature distribution. This frequency-aware adjustment mitigates over-smoothing while enhancing the propagation of discriminative information, particularly around edges or structured regions.

Importantly, our HGFE module is designed to be fully differentiable, allowing end-to-end training alongside existing CNN backbones without requiring separate graph construction or feature engineering steps. To validate its effectiveness, we conducted comprehensive experiments across multiple visual recognition benchmarks, including image classification (CIFAR-100 \cite{cifar100}), object detection (PASCAL VOC \cite{voc} and VisDrone \cite{visdrone}), and instance segmentation (CrackSeg \cite{crack} and CarParts \cite{carseg}). The consistent performance improvements demonstrate the broad applicability and robustness of our proposed framework.

The main contributions of this study are summarized as follows:
\begin{itemize}
	\item{We propose a unified framework that introduces hierarchical graph reasoning into CNNs, addressing their limitations in capturing non-local and topological dependencies.}
	\item{We design a dual-level graph convolution mechanism, combining intra-window and inter-window graphs to jointly model local spatial relations and global semantic context.}
	\item{We develop an AFM module to dynamically balance spectral components in graph propagation, enhancing feature fidelity and structural expressiveness.}
        \item{We conduct comprehensive experiments across diverse visual recognition tasks, showing that our approach yields consistent improvements over standard CNNs and prior graph-based baselines.}
\end{itemize}

The rest of this paper is organized as follows: Section \uppercase\expandafter{\romannumeral 2} reviews related works. Section \uppercase\expandafter{\romannumeral 3} details the proposed HGFE framework. Section \uppercase\expandafter{\romannumeral 4} presents the experimental results and analysis. Finally, Section \uppercase\expandafter{\romannumeral 5} concludes the paper and outlines future research directions.

\section{Related Work}
\subsection{CNNs in Context Modeling}
CNNs are widely regarded as powerful feature extractors because of their localized convolutional kernels and hierarchical architecture. Through weight sharing and spatial locality, CNNs can effectively learn low-level visual patterns, such as edges and textures, and progressively build up to higher-level abstractions. For a convolutional kernel of size \((2k+1)\times(2k+1)\), a single layer computes each output activation as a weighted sum over a local neighborhood, and stacking \(L\) such layers yields a theoretical effective receptive field (ERF) of approximately \((2kL+1)\times(2kL+1)\). However, empirical studies have revealed that the actual influence distribution within the ERF is heavily concentrated near the center, with peripheral pixels contributing significantly less to the output~\cite{relateA0}. This results in an inherent limitation: standard CNNs are inefficient at modeling long-range dependencies, particularly in tasks that require holistic scene understanding or global context reasoning. To alleviate the locality constraint without sacrificing spatial resolution, dilated convolutions were introduced~\cite{relateA1}. By introducing a dilation rate \(d\), the kernel samples the input features with spatial gaps, thereby enlarging the receptive field:
\begin{equation}
y_{i,j} = \sum_{u=-k}^{k} \sum_{v=-k}^{k} w_{u,v} \cdot x_{i+du,\, j+dv},
\end{equation}
where \((i,j)\) denotes the spatial position and \(w_{u,v}\) represents the convolution kernel weights. Although dilated convolutions are effective in extending contextual coverage for tasks like instance segmentation~\cite{relateA2}, they may introduce gridding artifacts and still lack explicit modeling of content-aware semantic relationships across distant regions.

To address this issue, non-local operations have been proposed~\cite{nonnet0}, which aggregate information globally by computing pairwise similarities between all spatial positions. Given a feature map \(x \in \mathbb{R}^{H \times W \times C}\), the non-local response at position \(i\) is computed as:
\begin{equation}
f(x)_i = \frac{1}{\mathcal{C}(x)} \sum_{j=1}^{HW} \phi(x_i)\, \psi(x_j)\, g(x_j),
\end{equation}
where \(\phi(\cdot)\) and \(\psi(\cdot)\) project the input into an affinity space, \(g(\cdot)\) transforms features, and \(\mathcal{C}(x)\) is a normalization factor (typically a softmax). This mechanism enables each position to attend to all others, thereby capturing global dependencies explicitly. However, its computational complexity is \(\mathcal{O}((HW)^2)\), making it inefficient for high-resolution inputs.

More recently, transformer-based architectures have been adapted to the visual domain, beginning with the ViT~\cite{vit}. These models divide an image into \(N\) patches and employ multi-head self-attention mechanisms to model pairwise dependencies:
\begin{equation}
\mathrm{Attention}(Q,K,V) = \mathrm{softmax}\left(\frac{QK^\top}{\sqrt{d}}\right)V,
\end{equation}
where \(Q, K, V \in \mathbb{R}^{N \times d}\) are the query, key, and value matrices, and \(d\) is the feature dimension. While ViT and its hierarchical extensions such as swin transformer and pyramid ViT achieve impressive performance by capturing long-range context~\cite{relateA3swintrains, relateA4pvt, relateA5swin2}, they suffer from quadratic complexity with respect to the number of patches (\(\mathcal{O}(N^2)\)) and typically require large-scale pretraining to converge effectively. Moreover, these architectures do not explicitly control the frequency characteristics of features, making it difficult to simultaneously preserve high-frequency local details (e.g., object boundaries) and low-frequency semantic context in a principled manner.

In summary, although CNNs can be enhanced via dilated convolutions, non-local blocks, or transformer modules to capture broader context, each of these strategies has limitations. Dilated convolutions remain fundamentally local, non-local methods incur high computational costs, and transformers demand large datasets and lack frequency-aware mechanisms. These observations highlight the need for a unified, hierarchical, and content-adaptive framework that can integrate both local and global context modeling while preserving critical structural details and maintaining computational efficiency.


\subsection{GSP and Spectral Filtering}

GSP extends classical signal processing techniques to irregular, non-Euclidean domains such as graphs, offering a principled framework to analyze and manipulate signals defined on graph-structured data~\cite{relateBgsp0, relateBgsp1, relateBgsp2, relateBgsp3}. In the context of GCNs, GSP provides the theoretical foundation for interpreting graph convolutions as spectral filtering operations, revealing their inherent frequency-selective behavior.
Let $\mathcal{G} = (\mathcal{V}, \mathcal{E})$ denote an undirected graph with $N = |\mathcal{V}|$ nodes, where the connectivity is encoded by an adjacency matrix $\mathbf{A} \in \mathbb{R}^{N \times N}$, and the degree matrix $\mathbf{D}$ is diagonal with entries $\mathbf{D}_{ii} = \sum_j \mathbf{A}_{ij}$. The normalized symmetric graph Laplacian is defined as:
\[
\mathbf{L}_{\mathrm{sym}} = \mathbf{I} - \mathbf{D}^{-\frac{1}{2}} \mathbf{A} \mathbf{D}^{-\frac{1}{2}},
\]
where $\mathbf{I}$ is the identity matrix. As a real symmetric positive semidefinite matrix, $\mathbf{L}_{\mathrm{sym}}$ admits an eigendecomposition:
\begin{equation}
\mathbf{L}_{\mathrm{sym}} = \mathbf{U} \mathbf{\Lambda} \mathbf{U}^\top,
\end{equation}
where $\mathbf{U} = [u_1, \dots, u_N]$ is an orthonormal matrix of eigenvectors (graph Fourier basis), and $\mathbf{\Lambda} = \mathrm{diag}(\lambda_1, \dots, \lambda_N)$ contains the corresponding non-negative eigenvalues, interpreted as graph frequencies. A graph signal $x \in \mathbb{R}^N$, defined as a feature vector over nodes, can be projected into the spectral domain via the graph Fourier transform: $\hat{x} = \mathbf{U}^\top x$. Its inverse is given by $x = \mathbf{U} \hat{x}$. Within this framework, a graph convolution is equivalent to a spectral filtering operation:
\begin{equation}
g_\theta(\mathbf{L}_{\mathrm{sym}})\,x = \mathbf{U}\,g_\theta(\mathbf{\Lambda})\,\mathbf{U}^\top\,x,
\end{equation}
where $g_\theta(\mathbf{\Lambda})$ denotes a learnable spectral filter function that selectively amplifies or attenuates specific frequency components, and $\theta$ are the parameters of the filter~\cite{gnn1}.

However, the direct computation of spectral convolution is computationally prohibitive for large-scale graphs, primarily due to the need for eigendecomposition of the normalized Laplacian matrix \(\mathbf{L}_{\mathrm{sym}}\). To alleviate this issue, polynomial approximation methods, such as Chebyshev polynomials, are commonly adopted to approximate the spectral filter \(g_\theta(\mathbf{\Lambda})\) without explicitly computing the eigenvector matrix \(\mathbf{U}\)~\cite{relateBgsp4chenet}. The $K$-order Chebyshev approximation takes the form:
\begin{equation}
g_\theta(\mathbf{L}_{\mathrm{sym}})\,x \approx \sum_{k=0}^{K} \theta_k\,T_k(\tilde{\mathbf{L}})\,x,
\end{equation}
where $T_k(\cdot)$ is the Chebyshev polynomial of degree $k$, and $\tilde{\mathbf{L}} = \frac{2}{\lambda_{\max}}\,\mathbf{L}_{\mathrm{sym}} - \mathbf{I}$ is the rescaled Laplacian. This formulation enables efficient, localized spectral filtering using sparse matrix operations.

Despite their computational efficiency, conventional spectral GCNs are prone to the well-documented issue of over-smoothing. As successive layers of low-pass graph filters are applied, node features tend to converge, resulting in the loss of discriminative information that is crucial for fine-grained visual recognition tasks. This phenomenon stems from the implicit suppression of high-frequency components, which are crucial for preserving local structures such as edges, boundaries, or isolated object parts. For visual understanding tasks, where both low-frequency global semantics (e.g., object categories, scene layout) and high-frequency local details (e.g., object contours, part-level distinctions) are vital, this frequency imbalance poses a significant limitation. Consequently, there is a growing need for adaptive spectral filtering mechanisms that can dynamically control the contribution of each frequency band based on task-specific signal characteristics.


In essence, understanding graph convolutions from the spectral perspective not only illuminates their functional behavior but also guides the principled design of frequency-aware, structure-preserving models. This insight forms the theoretical backbone of our HGFE architecture, which incorporates GSP principles into a visual pipeline that unifies global contextual reasoning with the preservation of local structural details.

\subsection{GCNs in Visual Tasks}
GCNs have emerged as a powerful paradigm for visual understanding tasks, owing to their capacity to model complex relational structures beyond the constraints of regular grid-based representations. Unlike traditional CNNs, which apply spatially shared filters over locally connected pixels, GCNs operate on arbitrarily structured graphs, allowing flexible message passing among nodes that encode semantic or spatial correlations. This flexibility enables GCNs to capture both short-range and long-range dependencies, making them particularly suited for tasks involving relational reasoning, such as scene graph generation, part-level segmentation, and contextual object understanding~\cite{relateC0, relateC1, relateC2, relateC3, relateC4, relateC5}.

However, effectively applying GCNs to vision tasks presents several unique challenges. Unlike domains where the graph structure is inherently defined (e.g., social or molecular networks), images lack an explicit relational graph, necessitating the construction of task-specific graphs to represent visual entities and their interactions. Common strategies include superpixel-based graphs, region proposal-based graphs, and feature map cell-level graphs~\cite{relateC6, relateC7, relateC8, relateC9}. Each approach involves a trade-off between representation granularity and computational cost. For instance, superpixels reduce graph complexity by aggregating homogeneous pixel regions but risk degrading detail due to imperfect segmentation. Region proposal-based graphs encode object-level relations but sacrifice finer spatial resolution. In contrast, cell-level or patch-level graphs preserve local detail but require deep GCN stacks to propagate information across distant nodes, increasing latency and susceptibility to over-smoothing.

To overcome the limitations of purely local message passing, several works have explored global reasoning mechanisms through learned or dense affinity graphs. For example, graph-based global reasoning networks~\cite{relateC10} projects CNN features into an interaction space to construct a fully connected graph and perform global context reasoning before projecting the enhanced features back. Learnable GCN~\cite{relateC11} learns pairwise affinities over feature maps to enable dense graph convolution for classification refinement. While these methods improve the capacity to model non-local dependencies, they incur substantial computational costs that typically scale quadratically with the number of nodes, rendering them impractical for high-resolution inputs. Furthermore, once affinity weights are learned, they remain static across samples and do not dynamically adapt to content-specific variations, limiting their ability to preserve fine-grained details in diverse visual scenes.

A persistent bottleneck in deep GCN architectures is the over-smoothing effect. As layers of graph convolutions are stacked, the repeated aggregation across neighborhoods causes node representations to become increasingly similar, blurring semantic distinctions that are critical for fine-grained recognition. This issue is particularly problematic in dense prediction tasks such as instance segmentation, where preserving high-frequency information, including object boundaries, corners, and texture details, is essential for accurate output. Attempts to mitigate this issue include architectural enhancements like residual connections~\cite{relateC12} or topology modifications through graph rewiring~\cite{relateC13}. However, these strategies often fail to explicitly account for the spectral characteristics of graph signals, and thus lack the ability to selectively control how different frequency components are propagated or suppressed.

The above limitations highlight a key research gap: how to design a graph-based mechanism that is both structurally expressive and frequency-aware, capable of maintaining task-relevant local detail while enabling global reasoning. Addressing this, our proposed HGFE framework introduces a frequency-adaptive graph processing paradigm, where the graph structure and the message propagation are dynamically modulated according to the spectral content of the input signals. By explicitly regulating the contributions of low-frequency and high-frequency components through a learnable gating mechanism, our approach provides a principled solution to balance global contextualization with fine-grained feature preservation, achieving superior performance across classification, detection, and segmentation tasks.

\section{Proposed Method}
\subsection{Framework Overview}

While CNNs serve as powerful backbones for visual representation learning, their limited receptive fields and spatially invariant operations restrict their ability to capture complex topological dependencies and adapt to varying frequency components in feature space. To address these issues in a unified and efficient manner, we propose a HGFE module, designed to be flexibly integrated into standard CNN pipelines for improved structural modeling and frequency-aware feature representation.

As illustrated in Fig.~\ref{fig:framework}, HGFE adopts a hierarchical graph reasoning strategy that jointly incorporates both local and global relational modeling. Specifically, given an intermediate feature map $\mathbf{F} \in \mathbb{R}^{B \times C \times H \times W}$, we divide it into non-overlapping spatial windows of size $h \times w$. For each window, a local graph is constructed where nodes correspond to feature vectors and edges represent learned semantic affinities. This local graph enables the extraction of fine-grained structural patterns through attention-guided message passing.
\begin{figure*}[t]
	\centering
	\includegraphics[width=7.0in]{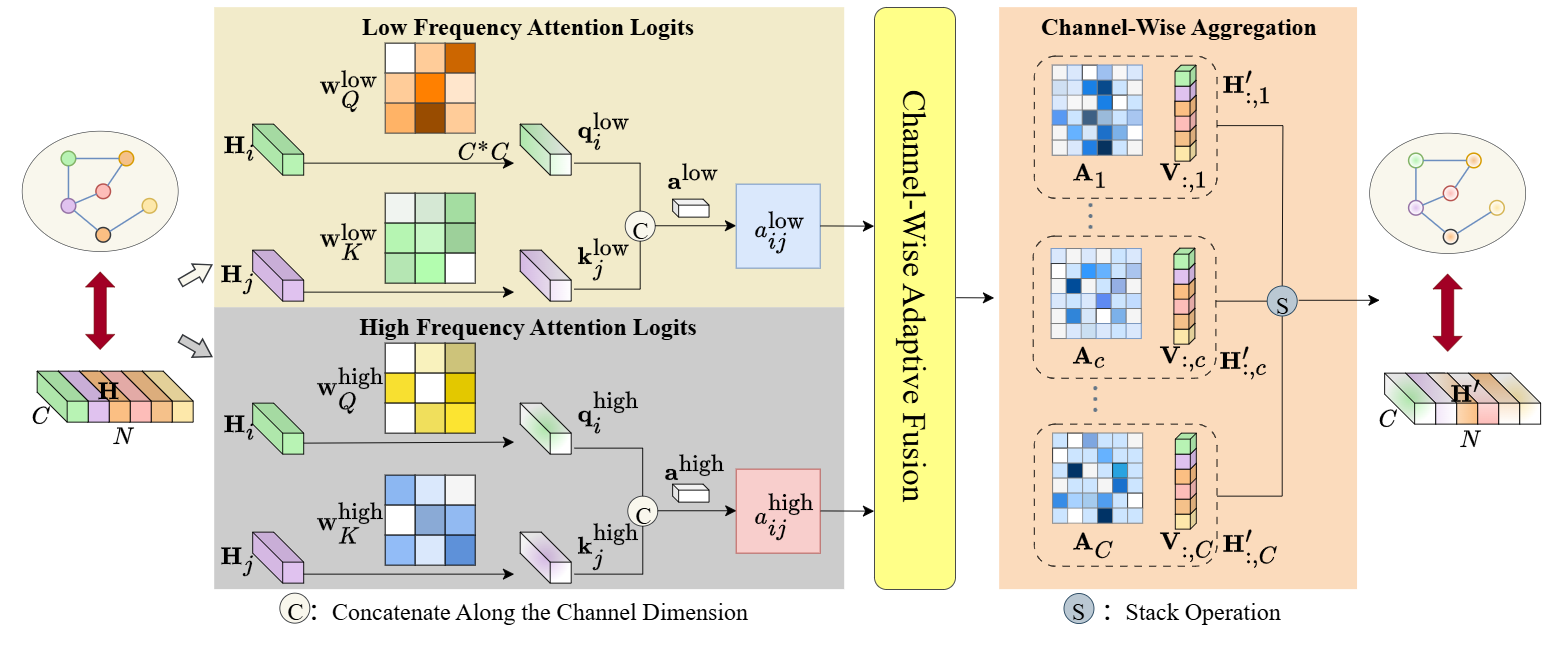}
	\caption{The proposed AFM module. AFM enhances feature learning by adaptively fusing low-frequency and high-frequency attention logits through a channel-wise gating mechanism. By leveraging both global context and fine-grained structure, AFM enables flexible and frequency-aware information aggregation.}
	\label{afm}
\end{figure*}

To capture broader contextual dependencies, we compress each window into a supernode via average pooling, forming a high-level graph over these representative nodes. A fully connected global graph is then constructed to enable efficient long-range information propagation. Importantly, both the local and global branches are enhanced by an AFM module, which decomposes node features into low-frequency and high-frequency components and dynamically fuses them using a channel-wise gating mechanism. This allows the model to emphasize edge-sensitive or context-sensitive information as needed, thereby mitigating over-smoothing and improving task-specific feature expressiveness.

Finally, the globally enhanced features are projected back to their corresponding spatial regions and fused with local representations via residual refinement. The result is a structurally enriched and frequency-adaptive feature map, which serves as a strong foundation for downstream tasks such as classification, detection, and segmentation.
\subsection{AFM}

In conventional GCNs, graph propagation typically relies on fixed low-pass spectral filters that encourage feature smoothing among neighboring nodes. While this aids in denoising, it often leads to over-smoothing, attenuating high-frequency components and impairing the discriminative power of learned representations. As discussed in Section~\uppercase\expandafter{\romannumeral 2}, standard first-order GCNs utilize fixed, low-order polynomial filters that substantially suppress high-frequency signals associated with large Laplacian eigenvalues \(\lambda_i\), resulting in uniform smoothing across the graph. However, this approach is ineffective for image-centric tasks because high-frequency information, such as object boundaries and fine textures, plays a crucial role in accurate representation.

To address this limitation, we introduce an AFM mechanism that dynamically adjusts the spectral response of each graph convolutional layer based on the input feature statistics. The overall architecture of AFM is shown in Fig.~\ref{afm}. Specifically, as illustrated in Algorithm~\ref{alg:afm}, given a node feature matrix \(\mathbf{H} \in \mathbb{R}^{N \times C}\), we first compute a global channel-wise summary vector:
\[
\mathbf{s} = \frac{1}{N} \sum_{i=1}^{N} \mathbf{H}_{i,:},
\]
which captures the mean activation per channel. This summary vector is then passed through a lightweight feedforward network and a sigmoid activation to generate a channel-wise gating vector:
\[
\boldsymbol{\alpha} = \sigma(\mathbf{W}_f \mathbf{s} + \mathbf{b}_f),
\]
where \(\mathbf{W}_f \in \mathbb{R}^{C \times C}\) and \(\mathbf{b}_f \in \mathbb{R}^C\) are learnable parameters. Each gate value \(\alpha_c\) interpolates between a low-pass and a high-pass spectral filter, yielding an adaptive frequency response:
\[
\theta_k^{(c)} = \alpha_c \theta_k^{\mathrm{low}} + (1 - \alpha_c) \theta_k^{\mathrm{high}}, \quad k = 0, \dots, K.
\]

In practice, we embed this modulation into graph attention networks (GAT)~\cite{gat3} by computing separate low-frequency and high-frequency attention logits, \(a_{ij}^{\mathrm{low}}\) and \(a_{ij}^{\mathrm{high}}\), and merging them via \(\alpha_c\):
\[
\tilde{a}_{ij}^{(c)} = \alpha_c \cdot a_{ij}^{\mathrm{low}} + (1 - \alpha_c) \cdot a_{ij}^{\mathrm{high}}.
\]
The final attention weights are normalized via softmax:
\[
a_{ij}^{(c)} = \frac{\exp\bigl(\sigma(\tilde{a}_{ij}^{(c)})\bigr)}{\sum_{j} \exp\bigl(\sigma(\tilde{a}_{ij}^{(c)})\bigr)},
\]
yielding a data-dependent mixture of low-frequency and high-frequency attention behaviors. Subsequently, node features are aggregated using learned value projections $\mathbf{V} = \mathbf{H} \mathbf{W}_V$, and updated as:
\[
\mathbf{H}'_{:,c} = \sigma(\mathbf{A}_c \mathbf{V}_{:,c}), \quad \text{for each channel } c = 1, \dots, C,
\]
where \(\mathbf{W}_V \in \mathbb{R}^{C \times C}\) is a learnable value projection matrix.

In summary, AFM enables each feature channel to adjust its frequency bias on the fly, facilitating a flexible balance between smoothing and detail preservation in graph-based feature learning.
\begin{algorithm}[t]
	\caption{AFM}
	\label{alg:afm}
	\begin{algorithmic}[1]
		\STATE \textbf{Input}: Node feature matrix $\mathbf{H} \in \mathbb{R}^{N\times C}$
		\STATE \textbf{Parameters}: 
		\begin{itemize}
			\item Low-pass attention: $\{\mathbf{W}_Q^{\mathrm{low}},\,\mathbf{W}_K^{\mathrm{low}},\,\mathbf{a}^{\mathrm{low}}\}$
			\item High-pass attention: $\{\mathbf{W}_Q^{\mathrm{high}},\,\mathbf{W}_K^{\mathrm{high}},\,\mathbf{a}^{\mathrm{high}}\}$
			\item Value projection: $\mathbf{W}_V \in \mathbb{R}^{C\times C}$
			\item Gating MLP: $\mathbf{W}_f \in \mathbb{R}^{C\times C},\ \mathbf{b}_f \in \mathbb{R}^C$
		\end{itemize}
		\STATE \textbf{Output}: Updated node features $\mathbf{H}' \in \mathbb{R}^{N\times C}$
		\vspace{0.3em}
		\STATE // Compute global channel-wise summary  
		\STATE $\mathbf{s} \gets \frac{1}{N} \sum_{i=1}^{N} \mathbf{H}_{i,:} \in \mathbb{R}^C$
		\vspace{0.3em}
		\STATE // Generate gating vector
		\STATE $\boldsymbol{\alpha} \gets \sigma\left( \mathbf{W}_f \mathbf{s} + \mathbf{b}_f \right) \in (0,1)^C$
		\vspace{0.3em}
		\STATE // Compute low-/high-frequency attention logits  
		\FOR{each node pair $(i,j)$}
		\STATE $\mathbf{q}_i^{\mathrm{low}} \gets \mathbf{H}_i \mathbf{W}_Q^{\mathrm{low}},\quad \mathbf{k}_j^{\mathrm{low}} \gets \mathbf{H}_j \mathbf{W}_K^{\mathrm{low}}$
		\STATE $a_{ij}^{\mathrm{low}} \gets \mathrm{LeakyReLU}\left( \left[ \mathbf{q}_i^{\mathrm{low}} \, \| \, \mathbf{k}_j^{\mathrm{low}} \right]^\top \mathbf{a}^{\mathrm{low}} \right)$
		\STATE $\mathbf{q}_i^{\mathrm{high}} \gets \mathbf{H}_i \mathbf{W}_Q^{\mathrm{high}},\quad \mathbf{k}_j^{\mathrm{high}} \gets \mathbf{H}_j \mathbf{W}_K^{\mathrm{high}}$
		\STATE $a_{ij}^{\mathrm{high}} \gets \mathrm{LeakyReLU}\left( \left[ \mathbf{q}_i^{\mathrm{high}} \, \| \, \mathbf{k}_j^{\mathrm{high}} \right]^\top \mathbf{a}^{\mathrm{high}} \right)$
		\ENDFOR
		\vspace{0.3em}
		\STATE // Channel-wise adaptive fusion of attention logits 
		\STATE $\tilde{a}_{ij}^{(c)}  \gets  \alpha_c \cdot a_{ij}^{\mathrm{low}} + (1 - \alpha_c) \cdot a_{ij}^{\mathrm{high}} $
		\vspace{0.3em}
		\STATE // Normalize attention coefficients  
		\STATE $a_{ij}^{(c)} \gets \mathrm{Softmax}_j\left( \tilde{a}_{ij}^{(c)} \right)$
		\vspace{0.3em}
		\STATE // Compute value representations and aggregate  
		\STATE $\mathbf{V} \gets \mathbf{H} \mathbf{W}_V$
		\STATE $\mathbf{H}'_{:,c} \gets \sigma\left(\mathbf{A}_c \mathbf{V}_{:,c} \right)$, where \(\mathbf{A}_c = [a_{ij}^{(c)}]\), \(c\) = 1,...,\(C\)
		\vspace{0.3em}
		\RETURN $\mathbf{H}'$
	\end{algorithmic}
\end{algorithm}

\subsection{Intra-Window Graph Convolution with AFM}

Traditional CNNs, which rely on fixed geometric local convolution operations, often struggle to effectively capture complex and irregular dependencies among pixels in non-Euclidean spatial domains, especially in regions containing edges, texture details, or irregular structures. To overcome this limitation, we introduce a graph convolution attention mechanism within each local window of the feature map, treating pixels inside the window as nodes of a fully connected graph. This allows explicit modeling of diverse and flexible interactions between nodes, thereby enhancing the representation capability for complex visual patterns.

As detailed in Algorithm~\ref{alg:intra_window_gat}, the input feature map \(\mathbf{F} \in \mathbb{R}^{B \times C \times H \times W}\) is partitioned into non-overlapping windows of size \(w \times w\), resulting in a total of \(B \cdot n_h \cdot n_w\) windows, where \(n_h = H / w\) and \(n_w = W / w\). Each window is regarded as a fully connected graph comprising \(N = w^2\) nodes, with each node corresponding to a spatial location and its associated feature vector.
For each window, the feature tensor is reshaped into a node feature matrix \(\mathbf{H} \in \mathbb{R}^{N \times C}\), followed by a dual-branch graph convolution structure designed to capture low-frequency, smooth semantic relations and high-frequency, detailed texture information respectively. These two frequency components are then adaptively fused via the AFM module, producing weighted and updated node features \(\mathbf{H}'\). Finally, the updated node features are then reshaped and aggregated to form the locally enhanced output feature map \(\mathbf{F}_{\mathrm{local}}\).

This mechanism enables the model to dynamically adjust the contribution of different frequency components, effectively capturing multi-scale local structural information and significantly enriching the expressiveness and discriminative power of the features. By enabling fine-grained local interactions and multi-frequency fusion, our method breaks through the fixed receptive field limitation of traditional convolutions, while enhancing sensitivity to texture details and edge information through a refined characterization of complex interactions among frequency components.
\begin{algorithm}[t]
\caption{Intra-Window Graph Convolution with AFM}
\label{alg:intra_window_gat}
\begin{algorithmic}[1]
\STATE \textbf{Input}: Feature map $\mathbf{F} \in \mathbb{R}^{B\times C\times H\times W}$, window size $w$
\STATE \textbf{Output}: Locally enhanced feature map $\mathbf{F}_{\mathrm{local}} \in \mathbb{R}^{B\times C\times H\times W}$
\STATE Partition $\mathbf{F}$ into non-overlapping windows:
\STATE \hspace{1em} $\{P_i \in \mathbb{R}^{C \times w \times w} \mid i = 1,\dots,B \cdot n_h \cdot n_w\}$

\FOR{each window $P_i$}
    \STATE Reshape $P_i$ into node features $\mathbf{H} \in \mathbb{R}^{N\times C}$, where $N = w^2$
    \STATE Apply AFM attention: $\mathbf{H}' \gets \mathrm{AFM}(\mathbf{H})$
    \STATE Reshape $\mathbf{H}'$ back to window feature $P_i' \in \mathbb{R}^{C \times w \times w}$
\ENDFOR
\STATE Reconstruct $\mathbf{F}_{\mathrm{local}}$ from all $P_i'$
\RETURN $\mathbf{F}_{\mathrm{local}}$
\end{algorithmic}
\end{algorithm}

\subsection{Inter-Window Super-Node Graph Convolution with AFM}

After local feature modeling within windows, we further construct a global representation by building an inter-window super-node graph to capture long-range dependencies across the feature map. Specifically, as illustrated in Algorithm~\ref{alg:inter_window_afm}, given the locally enhanced feature map \(\mathbf{F}_{\mathrm{local}} \in \mathbb{R}^{B \times C \times H \times W}\), where \(B\) denotes the batch size, we divide it into non-overlapping windows of size \(w \times w\), yielding a total of \(M = n_h \cdot n_w = \frac{H}{w} \cdot \frac{W}{w}\) windows. Each window is denoted by a tensor \(P'_i \in \mathbb{R}^{C \times w \times w}\).

To obtain a global semantic abstraction, each window is compressed into a super-node feature vector via spatial average pooling:
\[
v_i = \mathrm{AvgPool}(P'_i),\quad i=1,\dots,B \cdot M.
\]
The resulting super-node vectors are stacked to form a tensor \(\mathbf{V} \in \mathbb{R}^{B \times M \times C}\), where each sample contains a graph composed of \(M\) super-nodes, each representing one window.

To enable information exchange between super-nodes, we construct a graph convolution attention mechanism over these nodes. Furthermore, we integrate the AFM module to enhance the representation by modeling both low-frequency and high-frequency interactions. Specifically, the super-node feature tensor is passed through the AFM module:
\[
\mathbf{V}' = \mathrm{AFM}(\mathbf{V}).
\]
This module computes frequency-aware attention over low-pass and high-pass branches and adaptively fuses them using a learned gating mechanism, thus capturing multi-frequency patterns effectively.
We reshape \(\mathbf{V}'\) into a spatial grid of shape \(\mathbb{R}^{B \times n_h \times n_w \times C}\), such that each enhanced super-node feature corresponds to the spatial location of its original window. To inject the global semantic information back into the full-resolution space, we tile each vector \(v'_{b,h,w} = \mathbf{V}'[b,h,w,:] \in \mathbb{R}^C\) into its corresponding spatial window:
\[
G_{b,h,w} = \text{Tile}(v'_{b,h,w}) \in \mathbb{R}^{C \times w \times w},
\]
where \(\text{Tile}(\cdot)\) duplicates the vector across spatial dimensions. Each enhanced local window \(P'_i\) is then concatenated with its corresponding global context \(G_{b,h,w}\) along the channel dimension, producing a tensor of shape \(\mathbb{R}^{2C \times w \times w}\). A \(1 \times 1\) convolution is applied to reduce the channel dimension back to \(C\):
\[
P''_i = \mathrm{Conv}_{1\times1}([P'_i \,\Vert\, G_{b,h,w}]).
\]
Finally, all refined windows \(\{P''_i\}\) are reassembled in their original spatial order to reconstruct the output feature map:
\(
\mathbf{F} \in \mathbb{R}^{B \times C \times H \times W}.
\)

By constructing a graph over window-level super-nodes and employing frequency-aware attention, this module efficiently captures global semantic dependencies while maintaining low computational complexity. Unlike full attention on the entire \(H \times W\) feature grid with \(\mathcal{O}((HW)^2)\) complexity, our approach restricts attention to \(n_h \times n_w\) super-nodes, significantly enhancing scalability for high-resolution inputs. Integrated with the intra-window graph convolution enhanced by the AFM module, this inter-window super-node mechanism establishes a hierarchical dual-layer graph representation, bridging fine-grained local modeling and coarse global reasoning. This architecture delivers strong representational capacity, computational efficiency, and end-to-end trainability, providing a unified framework for structured and scalable visual feature learning.

\begin{algorithm}[t]
\caption{Inter‐Window Super‐Node Graph Convolution with AFM}
\label{alg:inter_window_afm}
\begin{algorithmic}[1]
\STATE \textbf{Input}: Feature map $\mathbf{F} \in \mathbb{R}^{B\times C\times H\times W}$, window size $w$
\STATE \textbf{Output}: Globally enhanced feature map $\mathbf{F}_{\mathrm{global}} \in \mathbb{R}^{B\times C\times H\times W}$
\STATE Partition $\mathbf{F}$ into non-overlapping windows:
\STATE \hspace{1em} $\{P_i \in \mathbb{R}^{C \times w \times w} \mid i = 1,\dots,B \cdot n_h \cdot n_w\}$
\FOR{each window $P_i$}
    \STATE $v_i \gets \mathrm{AvgPool}(P_i)\in\mathbb{R}^C$
\ENDFOR
\STATE Stack super‐nodes: $\mathbf{V}\in\mathbb{R}^{N\times C}$, where $N =B\times n_h\times n_w$
\STATE Apply AFM attention: $\mathbf{V}' \gets \mathrm{AFM}(\mathbf{V})$
\STATE Reshape $\mathbf{V}'$ to grid $\in\mathbb{R}^{B\times n_h\times n_w\times C}$
\FOR{each window index $(b,h,w)$}
    \STATE $G_{b,h,w} \gets \mathrm{Tile}(\mathbf{V}'[b,h,w,:]) \in \mathbb{R}^{C\times w\times w}$
    \STATE $P''_i \gets \mathrm{Conv}_{1\times1}([P_i \,\Vert\, G_{b,h,w}])$
\ENDFOR
\STATE Reconstruct $\mathbf{F}_{\mathrm{global}}$ from all $P''_i$
\RETURN $\mathbf{F}_{\mathrm{global}} $
\end{algorithmic}
\end{algorithm}

\subsection{Integration with CNN Backbone and Implementation Details}

The HGFE module integrates seamlessly after any intermediate CNN block without modifying the backbone architecture. Given an input feature tensor \(\mathbf{F} \in \mathbb{R}^{B \times C \times H \times W}\), it partitions the feature map into non-overlapping windows of fixed size \(w \times w\). Within each window, local graph convolution captures fine-grained spatial relationships, followed by average pooling to aggregate each window into a super-node. Global graph convolution then models long-range dependencies among these super-nodes. The AFM module adaptively fuses low-frequency and high-frequency components based on global statistics to balance detail preservation and noise suppression.

Regarding parameter complexity, the HGFE module introduces approximately \(4 C d + 3 C^{2} + C\) learnable parameters, where \(d\) denotes the embedding dimension. The computational complexity can be expressed as
\[
\mathcal{O}\bigl(B \cdot n_h \cdot n_w \cdot w^{2} (d + C + C^{2})\bigr) + \mathcal{O}\bigl(B \cdot (n_h n_w)^{2} \cdot d \bigr),
\]
where \(B\) is the batch size, \(C\) is the number of feature channels, \(H, W\) are the spatial dimensions of the input feature map, \(n_h = \lfloor H / w \rfloor\) and \(n_w = \lfloor W / w \rfloor\) denote the number of windows along the height and width dimensions respectively, and \(w\) is the fixed window size (height and width). Although the global graph convolution introduces quadratic complexity with respect to the number of super-nodes, the choice of window partitioning and embedding dimension ensures computational feasibility under typical settings. The entire HGFE module is fully differentiable and trained in an end-to-end manner, allowing straightforward integration into various CNN-based frameworks for classification, detection, and segmentation tasks. Experimental evaluations demonstrate the effectiveness of HGFE in enhancing feature representations and improving downstream task performance.

\section{Experiments}
In this section, we thoroughly assess the performance of the proposed HGFE framework through experiments designed for three key computer vision tasks: image classification, object detection, and instance segmentation. These tasks span coarse-grained category prediction to fine-grained structural understanding, covering a wide range of visual scenarios. All experiments were conducted using PyTorch and trained on an NVIDIA RTX A6000 GPU.
\begin{figure*}[t]
	\centering
	\includegraphics[width=\textwidth, height=0.4\textheight, keepaspectratio]{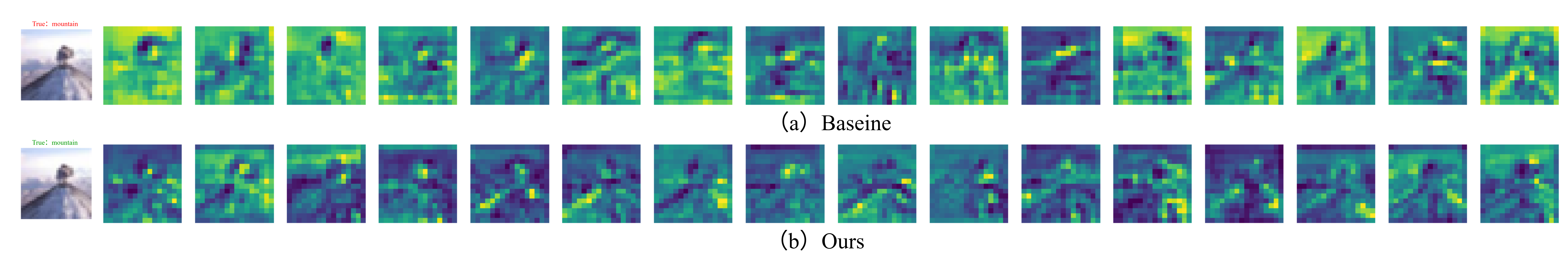}
	\caption{Comparison of feature activation patterns. (a) Baseline model shows diffuse activation; (b) Proposed model achieves activations along object contours across channels, enhancing geometric structure representation.}
	\label{fig:feature_activation}
\end{figure*}
\subsection{Experimental Setup}
\subsubsection{Datasets and Metrics}
To comprehensively evaluate the performance of the HGFE framework, we selected a series of representative datasets that encompassed a wide range of real-world complexities and structural variations across multiple vision tasks. For image classification, we used the CIFAR-100 dataset, which consists of 100 categories exhibiting substantial intra-class diversity. For object detection, experiments were conducted on the PASCAL VOC and VisDrone datasets, representing conventional detection scenarios and challenging aerial imagery, respectively. For instance segmentation, we employed a crack segmentation dataset and the CarParts dataset, covering diverse pixel-level prediction tasks ranging from sparse linear structures to densely annotated vehicle components. Regarding the evaluation metrics, we adopted standardized measures that are widely recognized within each task domain. The classification performance is reported using Top-1 and Top-5 accuracies. For detection, the mean average precision (mAP) at IoU thresholds of 0.5 and 0.5 to 0.95 were utilized. Instance segmentation performance is assessed via standard COCO metrics~\cite{coco}, including Box and Mask $\text{mAP}_{\text{0.5}}$ and $\text{mAP}_{\text{0.5:0.95}}$. These metrics comprehensively assess the model’s ability to localize and segment object instances across varying IoU thresholds.

\subsubsection{Model Selection}
To ensure fair and consistent evaluation across diverse visual tasks, we adopt the state-of-the-art YOLOv12~\cite{det3yolov12} framework as the unified baseline throughout all experiments. As a highly flexible and high-performance architecture, YOLOv12 is capable of addressing image classification, object detection, and instance segmentation within a single unified paradigm. This standardized setup enables a rigorous and comprehensive assessment of the proposed HGFE module, ensuring its effectiveness is evaluated under a representative and competitive visual learning framework. It is important to note that although YOLOv12 is our core baseline in this study, the proposed HGFE module is model-agnostic by design and can be flexibly incorporated into a variety of CNN-based or transformer-based architectures. We deliberately avoided using multiple disparate baseline methods to ensure a controlled evaluation environment and avoid confounding factors introduced by inconsistent architectural or training differences. Instead, we focus on rigorously comparing the HGFE-enhanced models with their baseline counterparts under consistent settings, highlighting the standalone effectiveness and adaptability of our feature enhancement strategy.

\subsubsection{Training Settings}
All models were trained using stochastic gradient descent (SGD) with a momentum of 0.937 and weight decay of $5 \times 10^{-4}$. The initial learning rate was uniformly set to 0.01 for all tasks, including classification, detection, and segmentation. A cosine annealing scheduler was adopted to gradually adjust the learning rate for smooth convergence. Additionally, a warm-up strategy was applied in the first 3 epochs, with a warm-up momentum of 0.8 and an initial bias learning rate of 0.1. Standard data augmentation techniques were employed during training, such as random horizontal flipping, resizing, cropping, and multi-scale input sampling, depending on the task. Unless otherwise specified, the window size $w$ in the intra-window attention of HGFE is set to 8 in all experiments.

\subsection{Image Classification}
To assess the classification performance of the proposed HGFE framework, we conducted experiments on the CIFAR-100 dataset. Owing to its relatively small image resolution ($32 \times 32$), we adopted a window size of $4$ in our graph construction for optimal local structure modeling. The YOLOv12 backbone was used as our baseline architecture, without any graph-based enhancement, to ensure a fair comparison.
\begin{table}[t]
\centering
\caption{Image classification performance comparison on CIFAR-100.}
\label{tab:cls-results}
\begin{tabular}{lcc}
\toprule
\textbf{Method} & \textbf{Top-1 Acc (\%)} & \textbf{Top-5 Acc (\%)} \\
\midrule
Baseline & 57.1 & 84.0 \\
Ours & \textbf{58.2} (\textcolor{green}{$\uparrow$\textbf{1.1}}) & \textbf{84.4} (\textcolor{green}{$\uparrow$\textbf{0.4}}) \\
\bottomrule
\end{tabular}
\end{table}
\begin{figure*}[t]
	\centering
	\includegraphics[width=\textwidth]{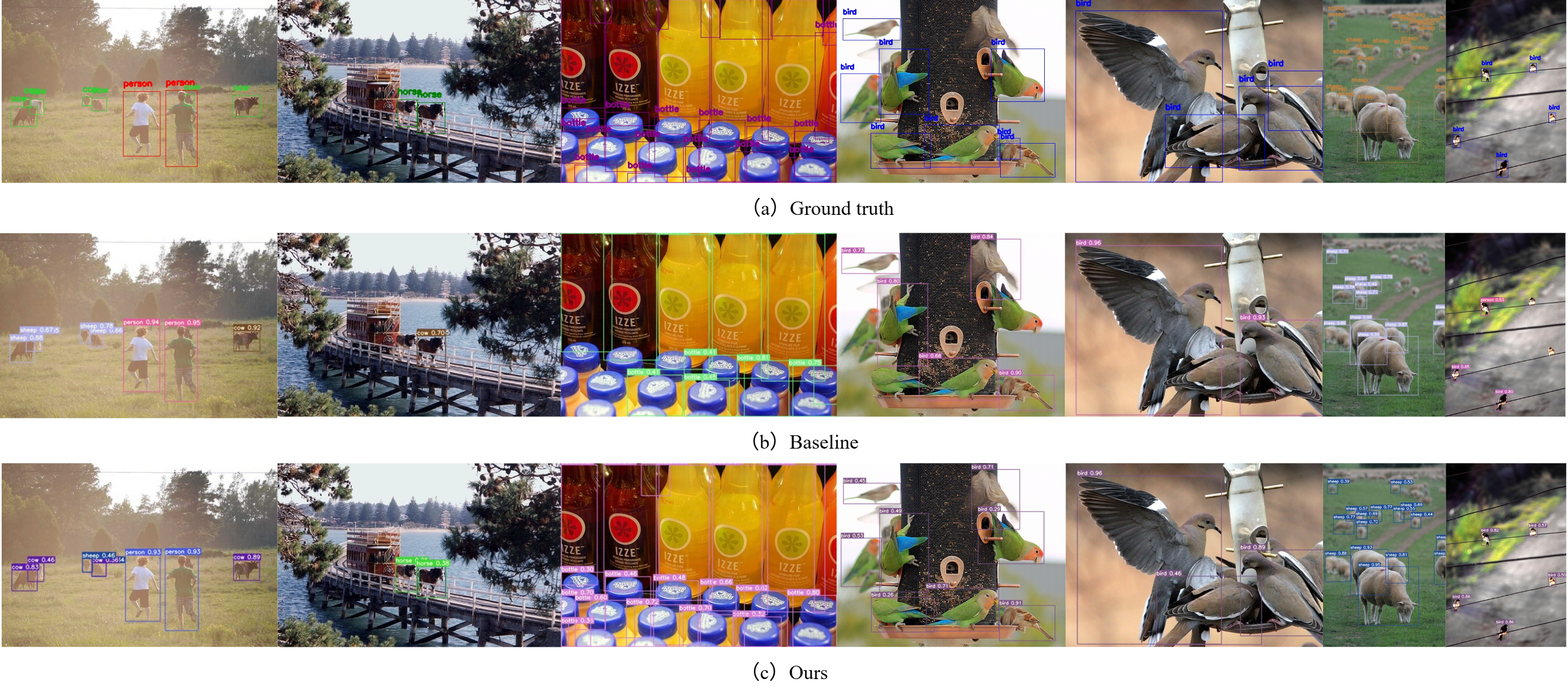}
	\caption{Qualitative visualization of detection results on the PASCAL VOC dataset. Each row from top to bottom corresponds to (a) Ground truth, (b) Baseline model predictions, and (c) Predictions from our HGFE-enhanced model.}
	\label{fig:detection_vis_voc}
\end{figure*}
\begin{figure*}[h!]
	\centering
	\includegraphics[width=\textwidth]{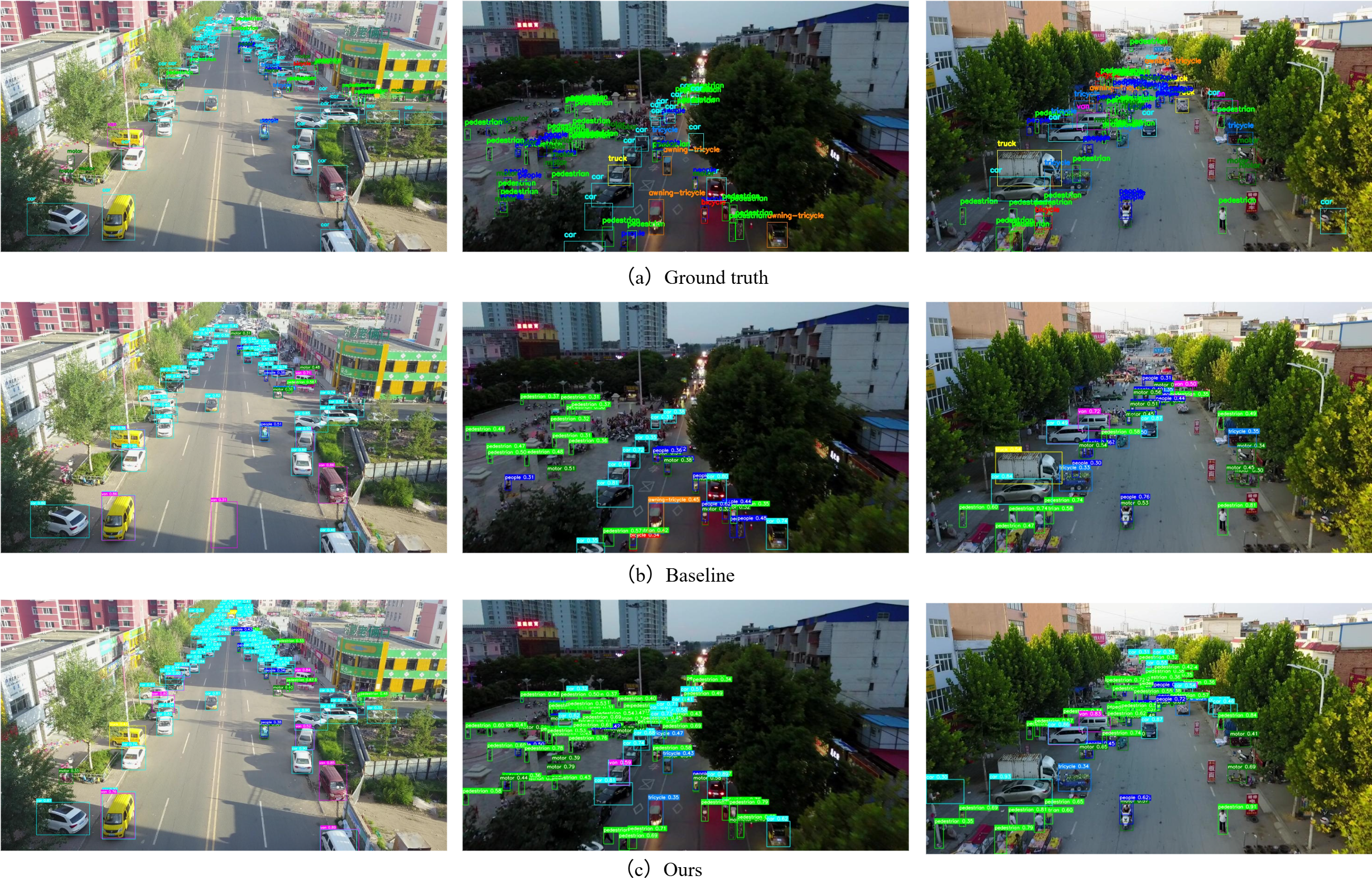}
	\caption{Qualitative visualization of detection results on the VisDrone dataset. Each row from top to bottom corresponds to (a) Ground truth, (b) Baseline model predictions, and (c) Predictions from our HGFE-enhanced model.}
	\label{fig:detection_vis_visdrone}
\end{figure*}

Table~\ref{tab:cls-results} presents the classification accuracy. Our method achieves a Top-1 accuracy of 58.2\%, improving over the baseline by 1.1\%, and a Top-5 accuracy of 84.4\%, with an absolute gain of 0.4\%. These gains confirm the advantages of frequency-adaptive graph learning for enhancing feature discrimination. 

To better illustrate the improvement brought by our model, we visualize the first 16 channel-wise feature maps for a representative test sample from the ``mountain'' class in Fig.~\ref{fig:feature_activation}. The green and red labels indicate correct and incorrect predictions, respectively. In the baseline model (a), the feature maps displayed scattered and noisy responses, failing to emphasize the structure of the object. The model misclassifies this sample because it lacks strong semantic cues. In contrast, our HGFE-enhanced model (b) exhibits significantly more structured and concentrated activation along the mountain contours. This spatial alignment is consistent across channels and reflects the enhanced capacity of the model to capture high-frequency features, such as edges and ridges, which are critical for differentiating complex natural scenes, such as mountains, from visually similar categories.
\begin{table}[t]
	\centering
	\caption{Object detection performance comparison on PASCAL VOC and VisDrone datasets}
	\label{tab:detection_results}
	\begin{tabular}{lccc}

		\toprule
		\textbf{Dataset} & \textbf{Method} & $\textbf{mAP}_{\textbf{0.5}}$ & $\textbf{mAP}_{\textbf{0.5:0.95}}$ \\
		\midrule
		\multirow{2}{*}{PASCAL VOC} 
		& Baseline & 84.5 & 62.6 \\
		& Ours & \textbf{85.2} (\textcolor{green}{$\uparrow$\textbf{0.7}}) & \textbf{62.9} (\textcolor{green}{$\uparrow$\textbf{0.3}}) \\
		\cmidrule{1-4}
		\multirow{2}{*}{VisDrone} 
		& Baseline & 49.5 & 28.5 \\
		& Ours & \textbf{50.7} (\textcolor{green}{$\uparrow$\textbf{1.2}}) & \textbf{29.1} (\textcolor{green}{$\uparrow$\textbf{0.6}}) \\
		\bottomrule
	\end{tabular}
\end{table}
\subsection{Object Detection}
To further validate the generalization capability of the proposed HGFE framework in dense prediction tasks, we evaluated it on two widely used object detection benchmarks: PASCAL VOC 2007+2012 and VisDrone-DET 2019. These datasets cover both conventional and complex real-world scenes, enabling a comprehensive evaluation under various visual conditions.

As shown in Table~\ref{tab:detection_results}, on the PASCAL VOC benchmark, our method achieved 85.2\% $\text{mAP}_{\text{0.5}}$ and 62.9\% $\text{mAP}_{\text{0.5:0.95}}$, surpassing the baseline by 0.7\% and 0.3\%, respectively. These consistent improvements highlight the effectiveness of our approach in enhancing feature representation for object detection in relatively clean and well annotated scenarios. On the more challenging VisDrone dataset, which involves dense scenes, small objects, and frequent occlusions, our method further demonstrates its robustness by achieving 50.7\% $\text{mAP}_{\text{0.5}}$ and 29.1\% $\text{mAP}_{\text{0.5:0.95}}$. This corresponds to improvements of 1.2\% and 0.6\% over the baseline, respectively. The results suggest that the proposed HGFE framework contributes positively to the model’s ability to capture spatial and geometric details that are critical for accurate object localization under complex real-world conditions.

To illustrate the effectiveness of our HGFE module more intuitively, we present a qualitative comparison of the detection results, as shown in Fig.~\ref{fig:detection_vis_voc} and \ref{fig:detection_vis_visdrone}. Examples were selected from both the VOC and VisDrone datasets to cover a range of scenarios, including crowded backgrounds, occlusions, and small-scale objects. In the baseline model, several objects were either mislocalized or entirely missed, particularly in visually cluttered or low-contrast regions. In contrast, our model demonstrated significantly improved localization accuracy and object completeness. Notably, our method more accurately identifies small or partially occluded instances, which are often overlooked by conventional detectors. This improvement can be attributed to the enhanced feature representations enabled by the HGFE framework, which effectively preserves the high-frequency details and contextual boundaries essential for precise object delineation.
\begin{figure*}[t]
	\centering
	\includegraphics[width=\textwidth]{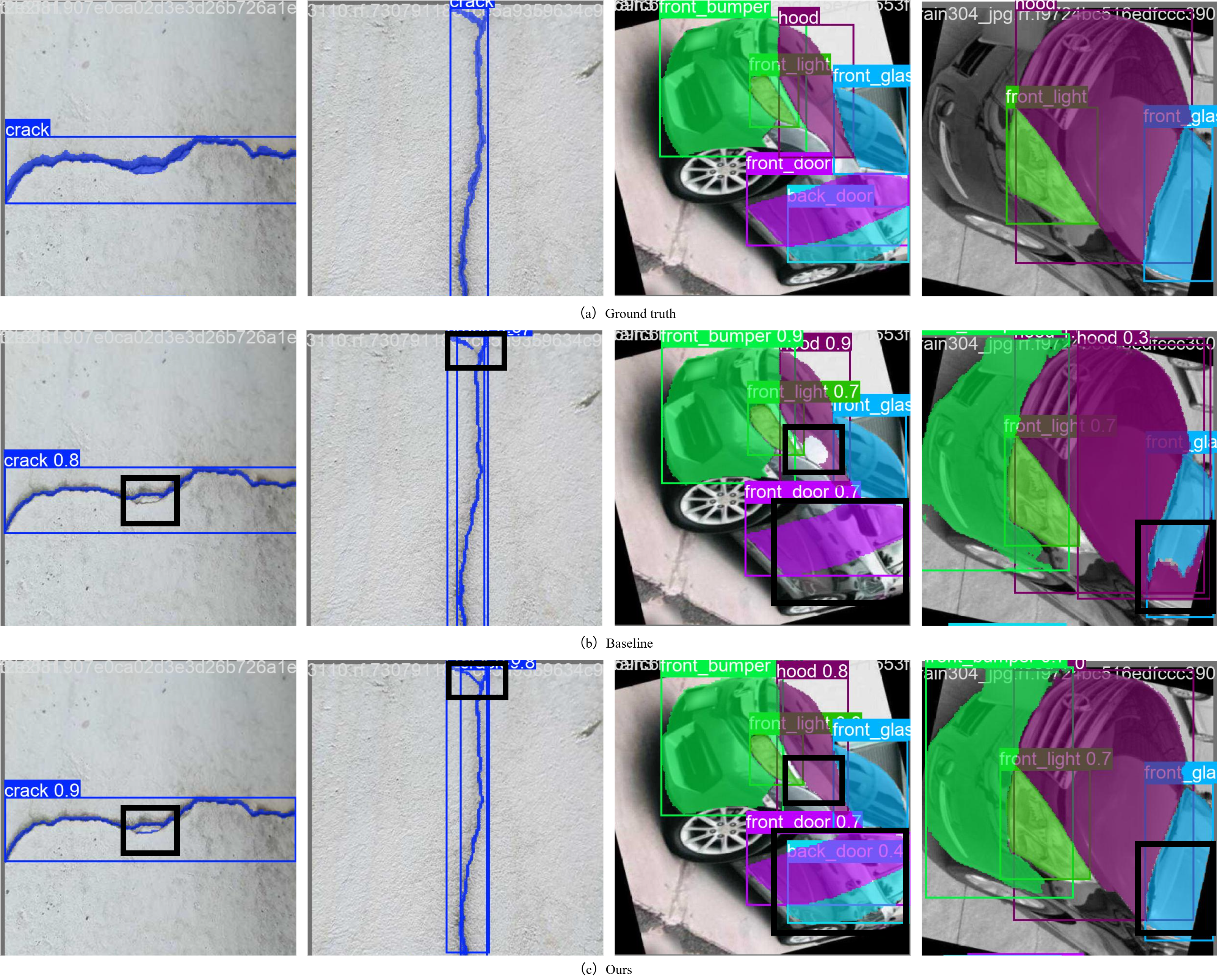}
	\caption{Qualitative comparison of instance segmentation results. Each row from top to bottom corresponds to (a) Ground truth, (b) Baseline model predictions, and (c) Predictions from our HGFE-enhanced model. The four columns represent different examples: the first two columns are from the crack segmentation dataset, and the last two columns are from the CarParts segmentation dataset.}
	\label{fig:Segmentation_vis}
\end{figure*}
\subsection{Instance Segmentation}
To examine the adaptability of the proposed HGFE framework for dense pixel-wise prediction tasks, we conducted instance segmentation experiments on two structurally distinct datasets: the crack segmentation dataset and the CarParts segmentation dataset. These datasets pose different levels of difficulty, ranging from detecting thin, elongated cracks to parsing fine-grained vehicle components, making them suitable for evaluating structural perception and contextual reasoning.
\begin{table}[t]
	\centering
	\caption{Instance segmentation performance comparison on Crack and CarParts datasets.}
	\label{tab:segmentation_results}
	\resizebox{\linewidth}{!}{ 
		\begin{tabular}{lccc}
			\toprule
			\textbf{Dataset} & \textbf{Method} & $\textbf{Box}_{\textbf{0.5}}$ / $\textbf{Box}_{\textbf{0.5:0.95}}$ & $\textbf{Mask}_{\textbf{0.5}}$ / $\textbf{Mask}_{\textbf{0.5:0.95}}$ \\
			\midrule
			\multirow{2}{*}{Crack} 
			& Baseline & 82.5 / 63.3 & 67.5 / 23.0 \\
			& Ours     & \textbf{83.1} / \textbf{63.7} (\textcolor{green}{$\uparrow$\textbf{0.6} / \textbf{0.4}}) & \textbf{68.4} / \textbf{23.3} (\textcolor{green}{$\uparrow$\textbf{0.9} / \textbf{0.3}}) \\
			\cmidrule{1-4}
			\multirow{2}{*}{CarParts} 
			& Baseline & 69.1 / 59.8 & 70.3 / 58.6 \\
			& Ours     & \textbf{69.9} / \textbf{60.3} (\textcolor{green}{$\uparrow$\textbf{0.8} / \textbf{0.5}}) & \textbf{71.6} / \textbf{59.2} (\textcolor{green}{$\uparrow$\textbf{1.3} / \textbf{0.6}}) \\
			\bottomrule
		\end{tabular}
	}
\end{table}

Table~\ref{tab:segmentation_results} reports the instance segmentation performance on the Crack and CarParts datasets. On the Crack dataset, our method achieves notable improvements over the baseline, with $\text{Box}_{\text{0.5}}$ increasing from 82.5\% to 83.1\% and $\text{Box}_{\text{0.5:0.95}}$ from 63.3\% to 63.7\%. Similarly, $\text{Mask}_{\text{0.5}}$ improves from 67.5\% to 68.4\% and $\text{Mask}_{\text{0.5:0.95}}$ from 23.0\% to 23.3\%. For the CarParts dataset, our model outperforms the baseline by +0.8\% and +0.5\% in box metrics, and +1.3\% and +0.6\% in mask metrics, respectively. These consistent gains across both datasets demonstrate the effectiveness of our proposed framework in enhancing instance-level recognition, particularly for objects with complex contours or subtle structural variations.

To supplement the numerical results, Fig.~\ref{fig:Segmentation_vis} offers a side-by-side visual comparison. Notably, baseline predictions often exhibit blurred boundaries and fragmented regions, particularly when dealing with narrow or irregularly shaped objects. In contrast, our model maintains sharper object contours and more coherent semantic regions. The observed improvement highlights HGFE’s ability of the HGFE to retain high-frequency edge details and propagate semantic context effectively, which is essential in both safety-critical infrastructure analysis and automotive perception tasks.

\subsection{Ablation Studies}
\begin{figure*}[t]
	\centering
	\includegraphics[width=\textwidth]{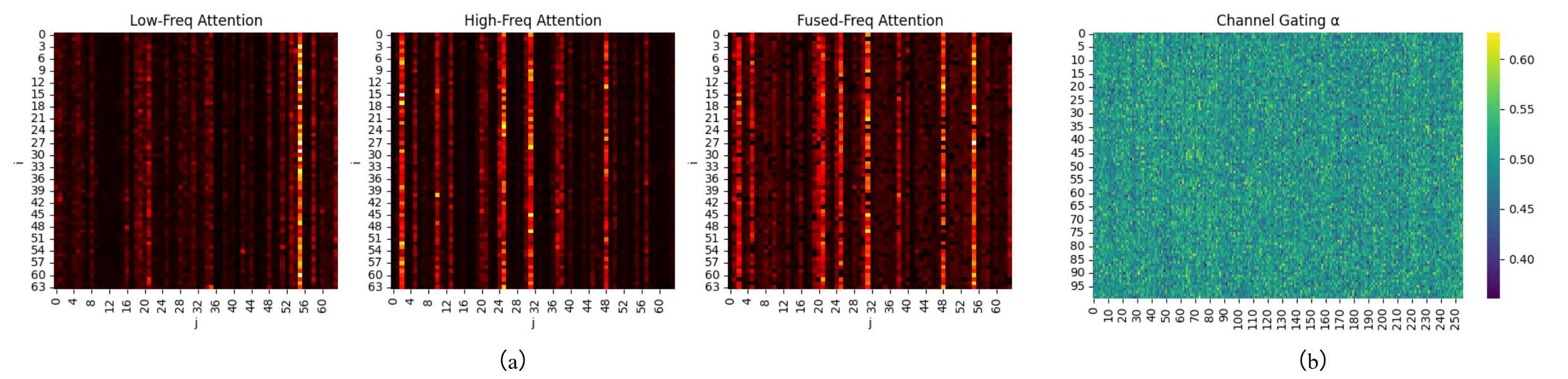}
	\caption{Visualization of AFM frequency responses. (a): spatial distributions of low-frequency and high-frequency attention after modulation. (b): learned channel-wise gating coefficients. AFM dynamically emphasizes frequency content based on semantic context.}
	\label{fig:afm_vis}
\end{figure*}
\subsubsection{Module Analysis}
To systematically assess the individual and combined contributions of the HGFE framework components, we conduct a stepwise ablation study based on the YOLOv12 baseline. As shown in Table~\ref{tab:ablation_modules}, we first integrate intra-window graph convolution and inter-window supernode interaction, both using fixed frequency filters. These modules provide consistent improvements by enhancing local structural reasoning and global contextual aggregation, respectively. Replacing the fixed filters with the proposed AFM, which employs learnable and input-dependent gating over frequency bands, further improves performance. This configuration achieves the highest accuracy, highlighting the essential role of AFM in dynamically adjusting frequency responses according to spatial and semantic variations. The results demonstrate that flexible and data-driven frequency routing is vital for enhancing the discriminative power of feature representations.

\begin{table}[t]
\centering
\caption{Ablation Study of Incremental Modules on VisDrone Dataset}
\label{tab:ablation_modules}
\begin{tabular}{lcc}
\toprule
\textbf{Model Variant} & $\textbf{mAP}_{\textbf{0.5}}$ & $\textbf{mAP}_{\textbf{0.5:0.95}}$ \\
\midrule
Baseline & 49.5 & 28.5 \\
+ Intra-Window Graph Conv  & 50.0 & 27.3 \\
+ Inter-Window Supernode    & 50.3 & 28.8 \\
+ AFM & \textbf{50.7} & \textbf{29.1}\\
\bottomrule
\end{tabular}
\end{table}

\begin{table}[t]
	\centering
	\caption{Ablation comparison of fixed frequency and adaptive AFM across multiple tasks.}
	\label{tab:afm_compare}
        \resizebox{\linewidth}{!}{ 
	\begin{tabular}{lcc}
		\toprule
		\textbf{Dataset / Task} & \textbf{Fixed Frequency(\%)} & \textbf{AFM (\%)} \\
		\midrule
		CIFAR-100 (Classification) & 57.8 (\text{Top-1 Acc}) & \textbf{58.2} \\
		PASCAL VOC (Detection) & 84.9 ($\text{mAP}_{\text{0.5}}$) & \textbf{85.2} \\
		VisDrone (Detection) & 50.3 ($\text{mAP}_{\text{0.5}}$) & \textbf{50.7} \\
		CrackSeg (Segmentation) & 68.0 ($\text{Mask}_{\text{0.5}}$) & \textbf{68.4} \\
		CarParts (Segmentation) & 71.1 ($\text{Mask}_{\text{0.5}}$) & \textbf{71.6} \\
		\bottomrule
	\end{tabular}
    }
\end{table}
\subsubsection{Frequency Modulation}

To assess the effectiveness of the proposed AFM mechanism, we perform an in-depth ablation study combining quantitative comparison with visual analysis. AFM enhances feature expressiveness by dynamically modulating low-frequency and high-frequency components during graph propagation. As shown in Fig.~\ref{fig:afm_vis}, the frequency response maps clearly demonstrate AFM’s selective amplification or suppression of different frequency bands. The learned channel-wise gating coefficients further reveal diverse activation patterns, indicating frequency-specific adaptation conditioned on input content. To quantify AFM’s contribution, we compare it with a fixed-frequency baseline that fuses frequency components without adaptive gating. As reported in Table~\ref{tab:afm_compare}, the fixed variant consistently underperforms across classification, detection, and segmentation tasks, underscoring the limitations of rigid frequency fusion. In contrast, AFM enables spatially and semantically adaptive frequency modulation, resulting in more robust, informative, and context-aware feature representations.
\begin{table}[t]
\centering
\caption{EImpact of Window Size $w$ on HGFE Performance: Detection Results on VisDrone Dataset}
\label{tab:window_size}
\begin{tabular}{cccc}
\toprule
\textbf{Window Size ($w$)} & $\textbf{mAP}_{\textbf{0.5}}$ & $\textbf{mAP}_{\textbf{0.5:0.95}}$ & \textbf{FLOPs (G)} \\
\midrule
4   & 49.9 & 28.6 & 96.0 \\
6   & 50.1 & 28.8 & 96.4 \\
8   & 50.7 & \textbf{29.1} & 103.5 \\
10  & \textbf{50.9} & \textbf{29.1} & 104.2 \\
12  & 50.3 & 28.9 & 107.3 \\
\bottomrule
\end{tabular}
\end{table}
\subsubsection{Window Size Sensitivity}
To evaluate the impact of window size $w$ in the HGFE framework, we examine its influence on both detection accuracy and computational cost. The window size governs the scope of intra-window graph convolution and frequency modulation, playing a critical role in balancing local structure modeling and global context aggregation. A small $w$ may lead to limited receptive fields and underutilization of context, whereas a large $w$ can dilute local spatial cues and increase computational overhead. As shown in Table~\ref{tab:window_size}, we assess a range of window sizes ${4, 6, 8, 10, 12}$ on the VisDrone detection benchmark, using YOLOv12 as the backbone. Performance improves consistently as $w$ increases from 4 to 8, with both $\text{mAP}_{\text{0.5}}$ and $\text{mAP}_{\text{0.5:0.95}}$ peaking around $w = 8$ and $w = 10$. Beyond this point, the accuracy gains begin to plateau or slightly drop, while FLOPs continue to grow. These results suggest that $w = 8$ offers an optimal trade-off between representational capacity and computational efficiency, effectively capturing both local and contextual features without incurring excessive complexity.

\subsection{Discussion}
The experimental results across the classification, detection, and segmentation tasks indicate that the proposed HGFE framework can effectively enhance feature representations for various visual tasks. The integration of intra-window graph convolution, inter-window supernode interaction, and AFM contributes to improved local structure modeling and better management of contextual dependencies. Ablation studies further supported the complementary role of each module. In particular, the frequency modulation mechanism helps the network adjust its focus on different frequency components, which may be beneficial for tasks with varying structural densities. The consistent performance improvements observed across different datasets suggest that the proposed enhancements are not overly sensitive to specific data domains or task types. By embedding the HGFE into a unified baseline and maintaining consistent settings, we can isolate and validate the contribution of each component. In summary, HGFE provides a simple yet effective enhancement to conventional feature extractors and may offer potential value when extended to other vision architectures or tasks.

\section{Conclusion}
This study presents a HGFE framework that addresses the critical limitations of conventional CNN architectures by introducing topologically aware, frequency-adaptive feature processing. Through a unified dual-graph design, the HGFE captures local spatial dependencies via intra-window graph convolutions while modeling long-range contextual interactions through inter-window supernode communication. The proposed AFM module further enables dynamic control over spectral responses, allowing each feature channel to selectively emphasize structural detail or contextual smoothness depending on its semantic characteristics. This fine-grained adaptability enhances the model’s ability to distinguish complex visual patterns without introducing an excessive computational overhead. When integrated seamlessly into standard CNN backbones, the HGFE demonstrates consistent performance improvements across multiple visual recognition tasks, including image classification, object detection, and instance segmentation. These results highlight the benefits of incorporating graph-based reasoning and frequency-aware modulation into convolutional frameworks. In future work, we plan to investigate HGFE’s applicability of HGFE in spatiotemporal domains and explore its synergy with transformer-based architectures to further enrich the structural modeling capacity.

\bibliographystyle{IEEEtran}
\bibliography{IEEEabrv,ref}
\end{document}